\documentclass[10pt,twocolumn,letterpaper]{article}
\usepackage{tikz}
\usepackage[pagenumbers]{cvpr} 

\definecolor{cvprblue}{rgb}{0.21,0.49,0.74}

\usepackage[pagebackref,breaklinks,colorlinks,allcolors=cvprblue]{hyperref}

\usepackage{makecell}

\usepackage{multirow}
\usepackage{makecell}

\usepackage{color}

\usepackage{multibib}
\newcites{supp}{References} 

\def\paperID{} 
\def\confName{CVPR}
\def\confYear{2026}

\title{Spatio-Temporal Difference Guided Motion Deblurring \\ with the Complementary Vision Sensor}

\author{
Yapeng Meng$^{1,\dagger}$, Lin Yang$^{3,\dagger,\ddagger}$, Yuguo Chen$^{1}$, Xiangru Chen$^{1}$, Taoyi Wang$^{4}$, Lijian Wang$^{1}$, \\
Zheyu Yang$^{4}$, Yihan Lin$^{2,*}$, Rong Zhao$^{1,*}$ \\
$^{1}$Department of Precision Instrument, Tsinghua University, Beijing, China, \\
$^{2}$Pen-Tung Sah Institute of Micro-Nano Science and Technology, Xiamen University, Xiamen, China, \\
$^{3}$Communication University of China, Beijing, China,~~$^{4}$Primevision Technology, Shanghai, China \\
{\tt\small
myp23@mails.tsinghua.edu.cn, yanglin2004@cuc.edu.cn, linyh@xmu.edu.cn,r\_zhao@tsinghua.edu.cn
}
}

\begin{document}
\twocolumn[{%
\renewcommand\twocolumn[1][]{#1}%
\maketitle
\centering
\vspace{-2em}
\includegraphics[width=\linewidth]{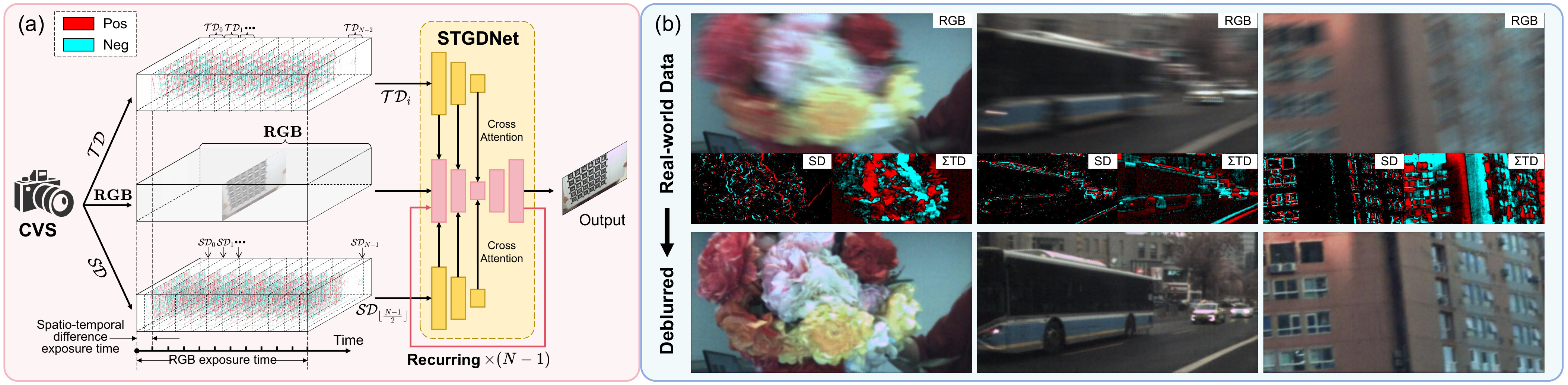}
\captionsetup{width=\linewidth}
\captionof{figure}{(a) Illustration of our deblurring framework and the Complementary Vision Sensor (CVS)~\cite{Tianmouc} data format. Within a single RGB exposure, the CVS simultaneously records high-frame-rate spatial difference ($\mathcal{SD}$) and temporal difference ($\mathcal{TD}$) signals, providing fine-grained structure and motion cues that guide the restoration of sharp content. Here, $\Sigma\mathcal{TD}$ denotes the visualization obtained by accumulating all $\mathcal{TD}$ signals within the exposure duration. (b) Our method achieves high-quality results on real-world captured scenes.}

\label{fig:1}
\vspace{1em}
}]
\begin{tikzpicture}[remember picture, overlay]
  \node [anchor=north, color=gray, align=center, font=\small] 
  at ([yshift=-0.1cm]current page.north) {
    This paper has been accepted for publication at the \\
    IEEE Conference on Computer Vision and Pattern Recognition (CVPR), Denver, 2026 \copyright IEEE
  };
\end{tikzpicture}

\def\thefootnote{$\dagger$}\footnotetext{These authors contributed equally to this work.}
\def\thefootnote{$\ddagger$}\footnotetext{This work is done during Lin Yang’s internship at Tsinghua.}
\def\thefootnote{*}\footnotetext{Corresponding author.}

\begin{abstract}


Motion blur arises when rapid scene changes occur during the exposure period, collapsing rich intra-exposure motion into a single RGB frame. Without explicit structural or temporal cues, RGB-only deblurring is highly ill-posed and often fails under extreme motion.
Inspired by the human visual system, brain-inspired vision sensors introduce temporally dense information to alleviate this problem. However, event cameras still suffer from event rate saturation under rapid motion, while the event modality entangles edge features and motion cues, which limits their effectiveness.
As a recent breakthrough, the complementary vision sensor (CVS), Tianmouc, captures synchronized RGB frames together with high-frame-rate, multi-bit spatial difference ($\mathcal{SD}$, encoding structural edges) and temporal difference ($\mathcal{TD}$, encoding motion cues) data within a single RGB exposure, offering a promising solution for RGB deblurring under extreme dynamic scenes. To fully leverage these complementary modalities, we propose \textbf{S}patio-\textbf{T}emporal Difference \textbf{G}uided \textbf{D}eblur \textbf{N}et (STGDNet), which adopts a recurrent multi-branch architecture that iteratively encodes and fuses $\mathcal{SD}$ and $\mathcal{TD}$ sequences to restore structure and color details lost in blurry RGB inputs. Our method outperforms current RGB or event-based approaches in both synthetic CVS dataset and real-world evaluations. Moreover, STGDNet exhibits strong generalization capability across over 100 extreme real-world scenarios. Project page: \url{https://tmcDeblur.github.io/}.
\end{abstract}


\section{Introduction}
\label{sec:intro}


High-speed motion during exposure causes severe blur because the sensor integrates several scene moments into a single RGB frame. Traditional deblurring methods—ranging from early kernel-based approaches~\cite{bahat2017reblurring,fergus2006removing,levin2011efficient,xu2013unnatural} to modern deep networks~\cite{chen2021hinet,nah2017deep,tsai2022banet,zamir2021multi}—struggle under such extreme blur. Large non-linear motion mixes structures and colors across the exposure, while the RGB modality lacks sufficient structural and motion cues to model the intra-exposure dynamics, making accurate deblurring fundamentally difficult.

To overcome this fundamental constraint, recent studies have explored leveraging additional visual modalities—particularly brain-inspired vision sensors known for their high temporal resolution—to assist RGB deblurring, such as using event cameras~\cite{DVS, Davis} or spiking cameras~\cite{vidar}. By fusing high temporal resolution data with RGB images, these methods aim to enhance motion awareness and improve deblurring under challenging conditions. 
However, the use of event cameras for deblurring still faces three main limitations.
(1) From the perspective of raw signal quality, event data are prone to false negatives during the refractory period~\cite{baldwin2020event, xu2021motion, dvs_refractory_time}, non-constant trigger thresholds~\cite{stoffregen2020reducing}, and saturation under rapid motion~\cite{gallego2020event}, all of which lead to degraded data fidelity. 
(2) From the modality perspective, Zhu et al.~\cite{zhu2025separation} point out that the event modality contains two entangled types of information—edge features and motion cues—which need to be disentangled in subsequent algorithms.
(3) From the hardware perspective, accurate spatio-temporal alignment between the event camera and RGB sensor typically requires complex optical setups and calibration procedures (e.g., beam splitters~\cite{han2020neuromorphic, kim2024cmta, duan2025eventaid}), limiting their practicality in real-world deployment.
Some recent advancements~\cite{Davis,kodama20231,guo20233} integrate event and intensity modalities within a single sensor, offering more convenient hardware support. However, they do not resolve the limitation of the event modality as mentioned above.

In this work, we employ a novel Complementary Vision Sensor (CVS), Tianmouc~\cite{Tianmouc}, which integrates two synergistic vision pathways (as shown in Fig.~\ref{fig:1}). The cognition-oriented pathway outputs 30 FPS RGB frames, while the action-oriented pathway captures ultra-high-speed (757–10,000 FPS) spatial difference ($\mathcal{SD}$) and temporal difference ($\mathcal{TD}$) signals. 
Benefiting from its fixed temporal rate with fixed multi-bit precision, the CVS has a bounded readout bandwidth and avoids saturation.
During long exposures or rapid scene motion, RGB frames inevitably suffer from motion blur, yet still preserve rich color and semantic information. 
In contrast, the $\mathcal{SD}$ and $\mathcal{TD}$ signals are captured with extremely short exposure durations, thus free from motion blur. Moreover, they respectively encode spatial structures and intra-exposure temporal dynamics, inherently decoupling edge features and motion cues at sensing level. 
By jointly leveraging these signals, the CVS achieves hardware-level spatio-temporal alignment across modalities, providing a foundation for high-fidelity deblurring.

Despite the advantages of CVS, several challenges remain for effective motion deblurring: the inconsistent RGB exposure time, the sparsity and lack of color in spatio-temporal difference data, and and the domain gap inherent in multi-modal fusion. To address this, we propose the \textbf{S}patio-\textbf{T}emporal Difference \textbf{G}uided \textbf{D}eblur \textbf{N}et (STGDNet).
Specifically, STGDNet adopts a multi-branch architecture that independently encodes and adaptively fuses the RGB frame, the sequence of $\mathcal{TD}$ data captured within the RGB exposure period, and the $\mathcal{SD}$ data captured near the RGB exposure midpoint. Since $\mathcal{TD}$ and $\mathcal{SD}$ encode luminance difference without color information, we integrate them with the RGB branch through an attention-based cross-modal fusion mechanism. To effectively model temporal dynamics and adapt to varying numbers of $\mathcal{TD}$, STGDNet employs a recurrent encoder-decoder design that sequentially processes each $\mathcal{TD}_i$ slice in temporal order and uses the $\mathcal{SD}$ feature to reinforce edge and texture details. At each iteration, the network refines the intermediate prediction via residual correction. This design ensures a balanced contribution from all modalities and improves the recovery of sharp, color-consistent images under severe motion blur.



Another significant challenge lies in real-world generalization. Networks trained only on synthetic datasets often generalize poorly to real-world scenarios~\cite{zhang2022survey}. To obtain suitable training data, we adopt a vision chip characterization method inspired by~\cite{meng2025technicalreportdmdbasedcharacterization} that converts existing high-frame-rate RGB sequences into the CVS data format. By randomizing the RGB exposure time, this method produces data with varying levels of motion blur. It employs a Digital Micromirror Device (DMD) and a corresponding optical setup to achieve high-speed, pixel-wise light control and project the modulated illumination onto the CVS sensor. Comprehensive evaluations on the synthetic dataset and \textbf{over 100 real-world scenes} demonstrate that our method achieves state-of-the-art motion deblurring performance.


The contribution of our paper is summarized as follows:



(1) We propose STGDNet, a Spatio-Temporal Difference Guided Deblurring Network that jointly leverages RGB, $\mathcal{SD}$, and $\mathcal{TD}$ modalities. By recurrently injecting spatio-temporal differences into the RGB feature space, our method more effectively models complex motion dynamics.

(2) Our method delivers state-of-the-art performance and strong real-world generalization without fine-tuning, enabled by the effectiveness of our data manufacturing pipeline.

(3) We establish a comprehensive dataset and performance boundary evaluation for the novel CVS deblurring task, including multi-exposure-time training data, diverse real-world test scenes, and a standardized real-captured benchmark for quantitative assessment.





\section{Related Work}
\label{sec:related}

\begin{figure*}
    \centering{\includegraphics[width=0.9 \linewidth]{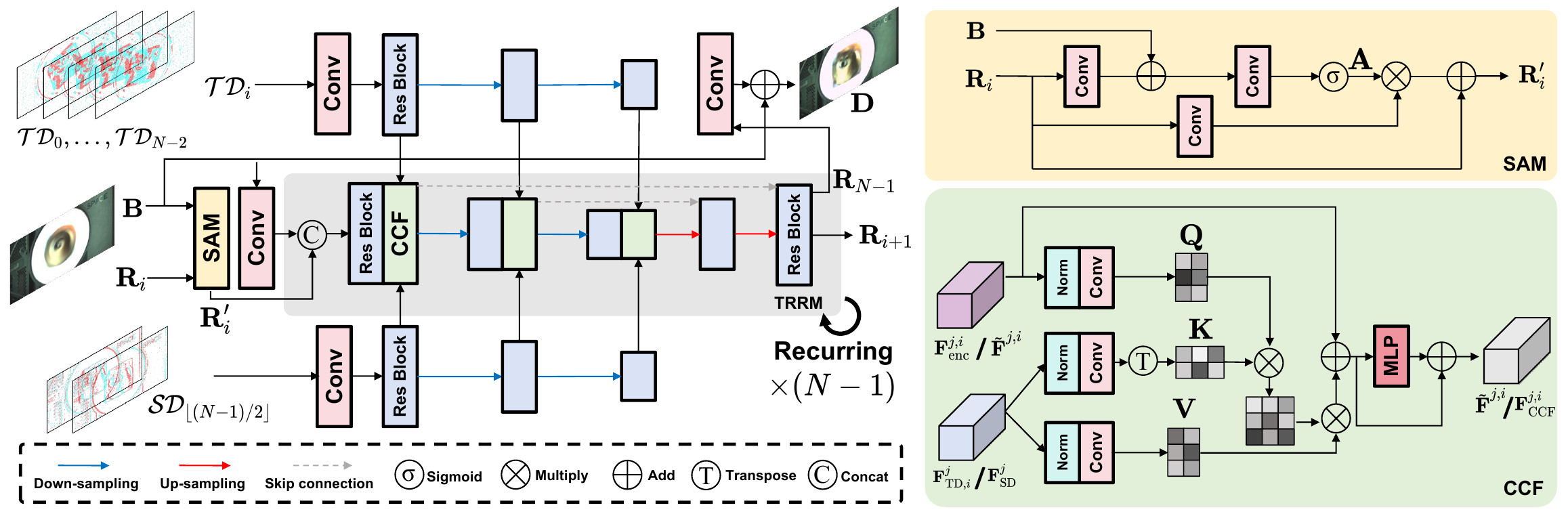}}\\
     \caption{Architecture of the Spatio-temporal Difference Guided Deblur Net (STGDNet), with including: Temporal Recurrent Refinement Module (TRRM), Supervised Attention Module (SAM), and Cross-modal Complementary Fusion (CCF).
     }
    \label{fig:2}
\end{figure*}

\paragraph{Image-based Motion Deblurring.}
Classical methods cast deblurring as inverse filtering with handcrafted priors and blur-kernel estimation, but struggle with spatially varying and complex motions~\cite{fergus2006removing,levin2011efficient}. Deep learning then shifted the paradigm to direct restoration, showing strong gains with encoder-decoder backbones, multi-scale and recurrent designs, and attention mechanisms~\cite{nah2017deep,tao2018scale,nah2019recurrent,zamir2021multi,zamir2022restormer}. Generative methods further improve perceptual quality~\cite{kupyn2018deblurgan}, and video deblurring aggregates neighboring frames to supply additional cues~\cite{zhu2022deep,pan2023deep}. However, existing image/video-based approaches still infer motion implicitly: the exposure-time motion trajectory is not directly observed, severe blur degrades alignment in video (multi-frame) deblurring and causes error accumulation. Consequently, in-the-wild extreme motion blur remains challenging without explicit motion priors.

\paragraph{Motion Deblurring with Brain-inspired Vision Sensors.}
Brain-inspired vision sensors provide high temporal resolution signals. Event-based methods evolved from physical models (e.g., Double Integral)~\cite{pan2019bringing} to end-to-end fusion networks~\cite{jiang2020learning,shang2021bringing} and better spatio-temporal interactions~\cite{sun2022eventfusion,Sun_2024_ECCV,yang2024stc}, with progress on cross-blur-scale generalization~\cite{yu2024scaleaware,zhang2023generalizing}, tackled spatio-temporal alignment~\cite{kim2024frequency,kim2024cmta}, low-light robustness~\cite{kim2024realevent,kim2025unified}, and unknown exposure~\cite{kim2022unknowntime}. Recently, Zhu et al.~\cite{zhu2025separation} revealed the inherent conflict between edge cues and motion cues in events and attempted to decouple them with algorithms. 

The complementary vision sensor (CVS)~\cite{Tianmouc}, provides a more complete data representation—capturing RGB frames together with temporal difference ($\mathcal{TD}$) and spatial difference ($\mathcal{SD}$) signals—along with multi-bit precision and the ability to avoid saturation under high-speed motion. Meng et al.~\cite{meng2025diffusion} proposed CBRDM, a diffusion-based model that leverages CVS signals to reconstruct high-speed and sharp RGB scenes in a unified framework. However, diffusion-based methods typically suffer from high computational cost, color distortion, and difficulties in maintaining the fidelity of the original scene.
In contrast, we develop a lightweight model which achieves more faithful color restoration and sharper structural recovery.
\section{Method}
\label{sec:method}

\subsection{Complementary Vision Sensor}
The CVS consists of two vision pathways: the RGB pathway ($\sim$30 FPS, 10-bit, $\mathbf{RGB} \in \mathbb{R}^{H \times W \times 3}$), and the high-speed spatio-temporal difference pathway (757–10,000 FPS, ±7-bit to ±1-bit, with a trade-off between frame rate and precision) that includes $\mathcal{SD} \in \mathbb{R}^{H \times W \times 2}$ and $\mathcal{TD} \in \mathbb{R}^{H \times W \times 1}$, which can be formally defined as:
\begin{equation}
\begin{split}
\mathcal{SD} &= \mathit{Concat}\left(\nabla_{+45^{\circ}} \mathbf{I} \;;\; \nabla_{-45^{\circ}} \mathbf{I}\right), \\
\mathcal{TD} &= \nabla_{t} \mathbf{I},
\end{split}
\end{equation}
where $\mathbf{I} \in \mathbb{R}^{H \times W}$ represents intensity, $\nabla_{\pm 45^{\circ}}$ denotes the spatial gradient computed along the $\pm 45^{\circ}$ directions (due to the cross-pixel architecture~\cite{Tianmouc} inside the CVS), $\mathit{Concat}(;)$ represents concatenation along the channel dimension, and $\nabla_t$ represents the temporal gradient.
In our deblurring experiments, the spatio-temporal difference pathway of the CVS operates at 757~FPS (corresponding to a $\tau_{\text{diff}} = 1{,}320$~µs interval between consecutive difference signals) with a precision of $\pm$7~bits.

\subsection{Problem Formulation} \label{sec:problem}

Our objective is to design an end-to-end deblurring model $\mathcal{M}$ that leverages the CVS data to reconstruct a sharp and color-fidelity image from a motion-blurred RGB frame.

As illustrated in Fig.~\ref{fig:1}(a), during a single RGB exposure period of duration $t_{\text{RGB}}$, the CVS simultaneously records:  (1) a sequence of $N$ spatial difference frames $\{\mathcal{SD}_k\}_{k=0}^{N-1}$; and (2) $N-1$ temporal difference frames $\{\mathcal{TD}_i\}_{i=0}^{N-2}$. 
Among the $\mathcal{SD}$ sequence, the middle frame $\mathcal{SD}_{\lfloor (N-1)/2 \rfloor}$, captured closest to the RGB exposure midpoint, provides stable structural and edge cues.  
The $\mathcal{TD}$ sequence, in contrast, encodes inter-frame motion dynamics, serving as an effective motion prior.  
These two complementary modalities are jointly exploited in our framework to guide deblurring from different perspectives.

Formally, the model $\mathcal{M}$ takes the blurred RGB frame $\mathbf{B}$, the central spatial difference frame $\mathcal{SD}_{\lfloor (N-1)/2 \rfloor}$, and all temporal difference frames $\{\mathcal{TD}_i\}_{i=0}^{N-2}$ as input, and outputs a restored sharp image $\mathbf{D}$:
\begin{equation}\label{eq:problem_formulation}
\mathbf{D} = \mathcal{M}\!\left(\mathbf{B},\, \mathcal{SD}_{\lfloor (N-1)/2 \rfloor},\, \{\mathcal{TD}_i\}_{i=0}^{N-2}\right),
\end{equation}
where $\mathbf{D}$ denotes the final deblurred result.  

The number of recorded difference frames $N$ is determined by the RGB exposure time $t_{\text{RGB}}$ and the CVS difference signal sampling interval $\tau_{\text{diff}}$:
\begin{equation}
N = \left\lceil \frac{t_{\text{RGB}}}{\tau_{\text{diff}}} \right\rceil,
\end{equation}
where $\lceil \cdot \rceil$ and $\lfloor \cdot \rfloor$ denote the ceiling and floor operations, respectively.  
The ceiling operation is applied because $t_{\text{RGB}}$ is not necessarily an integer multiple of the difference sampling interval $\tau_{\text{diff}}$, this ensures that all motion cues occurring within the RGB exposure are fully covered. Similarly, since $\mathcal{SD}$ frames are discretely sampled in time, the middle index $\lfloor (N{-}1)/2 \rfloor$ points to the frame that is temporally closest to the exposure midpoint.
We use only this middle $\mathcal{SD}$ frame---rather than the entire $\mathcal{SD}$ sequence---to enforce explicit structural alignment between the restored image and a physically captured structural snapshot. 

\subsection{Spatio-Temporal Difference Guided Deblurring Framework}
As illustrated in Fig.~\ref{fig:2}, our STGDNet is an encoder-decoder framework designed to recover sharp and color-fidelity images from blurred RGB inputs by leveraging the complementary spatio-temporal difference signals captured by the CVS. The framework consists of three key components: (1) modality-specific feature extraction branches for $\mathcal{SD}$, and $\mathcal{TD}_i$ inputs; (2) a Temporal Recurrent Refinement Model (TRRM) for dynamic temporal progressively deblur; and (3) a Cross-modal Complementary Fusion (CCF) module for progressive multimodal fusion. 

\paragraph{SD/TD Feature Encoding.}
The spatial- and temporal-difference inputs are encoded by two individual encoders composed of two $3{\times}3$ convolution layers with Leaky ReLU activations, residual connections, and optional strided downsampling. 
The $\mathcal{SD}$ encoder extracts structural edge cues from the mid-exposure frame, while the $\mathcal{TD}$ encoder captures motion dynamics from each $\mathcal{TD}_i$. 
Formally,
\begin{equation}
\mathbf{F}_{\text{SD}} = \mathcal{E}_{\text{SD}}(\mathcal{SD}_{\lfloor (N-1)/2 \rfloor}), 
\qquad
\mathbf{F}_{\text{TD}_i} = \mathcal{E}_{\text{TD}}(\mathcal{TD}_i),
\end{equation}
where $\mathcal{E}_{\text{SD}}$ and $\mathcal{E}_{\text{TD}}$ denote the SD and TD encoders, respectively. 
The resulting multi-scale features are projected via $1{\times}1$ convolutions for channel alignment and injected into the TRRM through CCF at corresponding scales.

\paragraph{Temporal Recurrent Refinement Module (TRRM).}
To effectively model long-exposure temporal motion cues, we design the TRRM, which progressively fuses the encoded temporal features $\{\mathbf{F}_{\text{TD}_i}\}$ and the spatial feature $\mathbf{F}_{\text{SD}}$ to refine the deblurring results. 

TRRM consists of multiple hierarchical encoder-decoder blocks, where each encoder stage performs spatio-temporal fusion via CCF, and each decoder stage reconstructs the deblurred feature map with skip connections for texture restoration:
\begin{equation}
\begin{split}
\mathbf{E}^{j+1} &= \mathcal{E}_{\text{TRRM}}^{j}\!\left(\mathbf{E}^{j},\, \mathbf{F}_{\text{TD}_i}^{j},\, \mathbf{F}_{\text{SD}}^{j}\right), \\
\mathbf{G}^{j+1} &= \mathcal{D}_{\text{TRRM}}^{j}\!\left(\mathbf{G}^{j},\, \mathbf{E}^{j}\right),
\end{split}
\end{equation}
where $\mathbf{E}^{j}$ and $\mathbf{G}^{j}$ are the encoder and decoder features at stage $j$, respectively; $\mathcal{E}_{\text{TRRM}}^{j}$ and $\mathcal{D}_{\text{TRRM}}^{j}$ denote the encoder and decoder within TRRM.

At each recurrent step $i$ ($i = 0, 1, \dots, N-2$), the TRRM outputs an intermediate residual map $\mathbf{R}_i$, which is then refined by the Supervised Attention Model (SAM) before being fed back for the next iteration:
\begin{equation}
\begin{split}
\mathbf{R}_{i}^{\prime} &= \text{SAM}\!\left(\mathbf{R}_{i},\, \mathbf{B}\right), \\
\mathbf{R}_{i+1} &= \text{TRRM}\!\left(\mathbf{R}_{i}^{\prime},\, \mathbf{B}_{\text{enc}},\, \mathbf{F}_{\text{TD}_i},\, \mathbf{F}_{\text{SD}}\right),
\end{split}
\end{equation}
where $\mathbf{B}_{\text{enc}}$ denotes the shallow feature embedding obtained by applying a single $3{\times}3$ convolution to the blurred RGB frame $\mathbf{B}$. 
This recurrent mechanism enables progressive motion deblur refinement and continuous enhancement across the spatio-temporal difference sequence. 
After the final recurrent step, the last residual map $\mathbf{R}_{N-1}$ is refined by a $3{\times}3$ convolution $\text{Conv}_{\text{out}}$ to predict the residual output $\mathbf{R}_{\text{out}}$, and the final restored image is obtained via a residual connection:
\begin{equation}
\mathbf{D} = \mathbf{B} + \text{Conv}_{\text{out}}(\mathbf{R}_{N-1}).
\end{equation}
By design, TRRM adapts to different sequence lengths $N$ corresponding to various exposure durations, ensuring flexibility to real-world motion blur.

\paragraph{Supervised Attention Module (SAM).}

The SAM~\cite{zamir2021multi} adaptively modulates the residual representation $\mathbf{R}_{i}$ using spatial attention mechanism.  
It consists of three convolutional layers: $\mathcal{C}_1$ extracts features, $\mathcal{C}_2$ projects current residuals into RGB domain for alignment with $\mathbf{B}$, and $\mathcal{C}_3$ generates a spatial attention map $\mathbf{A}$.
Formally,
\begin{equation}
\begin{split}
\mathbf{A} &= \sigma(\mathcal{C}_3(\mathcal{C}_2(\mathbf{R}_i) + \mathbf{B})), \\ 
\mathbf{R}_{i}' &= \mathbf{R}_{i} + \mathcal{C}_1(\mathbf{R}_i) \odot \mathbf{A}.
\end{split}
\end{equation}
where $\mathcal{C}_{1}$–$\mathcal{C}_{3}$ denote convolutional layers and $\sigma(\cdot)$ is the sigmoid function. This gate reinforces feature correlated with the blurred regions.

\paragraph{Cross-modal Complementary Fusion (CCF).}
To jointly exploit motion and structure cues from the TD and SD modalities, CCF is embedded in each TRRM encoder stage.  
At recurrent step $i$ and scale $j$, let $\mathbf{F}_{\text{enc}}^{j,i}$ denote the current encoder feature, $\mathbf{F}_{\text{TD},i}^{j}$ denote the TD feature at step $i$, and $\mathbf{F}_{\text{SD}}^{j}$ denote the corresponding SD feature.  
CCF consists of two cascaded attentional fusions: the first enhances motion awareness by integrating $\mathbf{F}_{\text{TD},i}^{j}$, and the second refines structure awareness using $\mathbf{F}_{\text{SD}}^{j}$.  
Formally,
\begin{equation}
\begin{split}
\tilde{\mathbf{F}}^{j,i} &=
\operatorname{softmax}\!\left(
\frac{(\mathbf{Q}_{\text{enc}}^{j,i})(\mathbf{K}_{\text{TD}}^{j,i})^{\!\top}}{\sqrt{d_k}}
\right)\mathbf{V}_{\text{TD}}^{j,i}
+ \mathbf{F}_{\text{enc}}^{j,i},\\
\mathbf{F}_{\text{CCF}}^{j,i} &=
\operatorname{softmax}\!\left(
\frac{(\tilde{\mathbf{Q}}^{j,i})(\mathbf{K}_{\text{SD}}^{j})^{\!\top}}{\sqrt{d_k}}
\right)\mathbf{V}_{\text{SD}}^{j}
+ \tilde{\mathbf{F}}^{j,i},
\end{split}
\end{equation}
where $\mathbf{Q}$, $\mathbf{K}$, and $\mathbf{V}$ are $1{\times}1$ convolutional projections applied to the corresponding feature maps, i.e.,
$\mathbf{Q}_{\text{enc}}^{j,i}{=}\phi_Q(\mathbf{F}_{\text{enc}}^{j,i})$,
$\mathbf{K}_{\text{TD}}^{j,i},\mathbf{V}_{\text{TD}}^{j,i}{=}\phi_{K,V}(\mathbf{F}_{\text{TD},i}^{j})$, and $\tilde{\mathbf{Q}}^{j,i}{=}\phi_Q(\tilde{\mathbf{F}}^{j,i})$,
$\mathbf{K}_{\text{SD}}^{j},\mathbf{V}_{\text{SD}}^{j}{=}\phi_{K,V}(\mathbf{F}_{\text{SD}}^{j})$.  
The first attention produces $\tilde{\mathbf{F}}^{j,i}$, a motion-enhanced middle representation, while the second yields $\mathbf{F}_{\text{CCF}}^{j,i}$, which incorporates both motion and structural cues.  
Embedding CCF across multiple scales allows TRRM to achieve hierarchical spatio--temporal feature aggregation.

\subsection{Loss Function}
We employ a PSNR-based loss between the predicted and ground-truth images, 
where $\lambda_{\text{psnr}} = 0.5$ and $\epsilon$ is a small constant to ensure numerical stability:
\begin{equation}
    \mathcal{L}_{\text{PSNR}} = - \lambda_{\text{psnr}} \cdot 10 \log_{10} \!\left( \frac{1}{\text{MSE} + \epsilon} \right).
\end{equation}

\subsection{Implementation Details}
All parameters are trained from scratch. The optimizer used is AdamW~\cite{loshchilov2019adamw}, with a learning rate set to $2 \times 10^{-4}$ and a weight decay of $1 \times 10^{-4}$. Momentum parameters ($\beta$) of AdamW are set to $[0.9, 0.99]$. The Cosine learning rate scheduler is applied with a minimum learning rate of $1 \times 10^{-7}$. We use 4 NVIDIA 4090 GPUs and train for 10 epochs.

\begin{figure*}[htbp]
    \centering{\includegraphics[width=1 \linewidth]{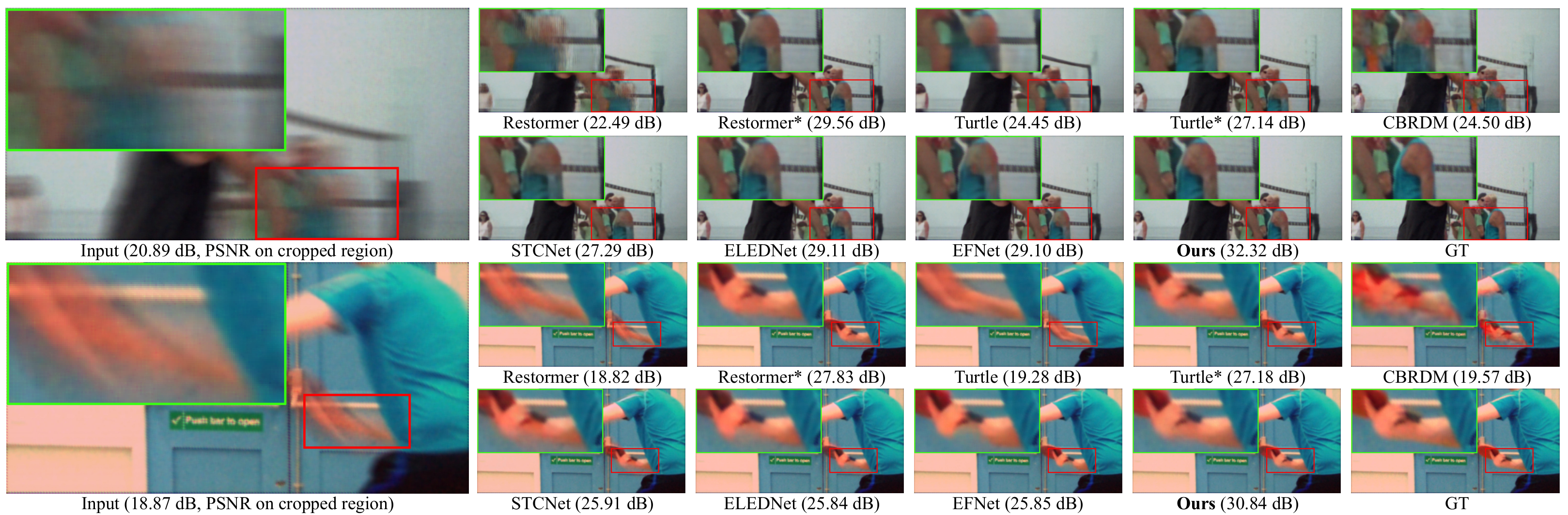}}\\
     \caption{Visualization of different methods on SportsSloMo-CVS dataset. PSNR values for the cropped regions are provided.}
    \label{fig:3}
\end{figure*}
\begin{table*}
\centering
\setlength{\tabcolsep}{4pt}  
\caption{
Comparison across different numbers of averaged blur frames on SportsSloMo-CVS dataset. 
The asterisk (*) denotes RGB-only methods augmented by concatenating $\mathcal{SD}$ and $\mathcal{TD}$ to the RGB input.
}
\begin{tabular}{c|c c|c c|c c|c c|c|c}
\toprule
\multirow{2}{*}{\rule{0pt}{2.6ex}Method} & \multicolumn{2}{c|}{5}  & \multicolumn{2}{c|}{7}   & \multicolumn{2}{c|}{9}   & \multicolumn{2}{c|}{11}   & \multirow{2}{*}{Params (M)$\downarrow$}  & \multirow{2}{*}{FLOPs (T)$\downarrow$}\\

\cline{2-9}

& PSNR$\uparrow$ & SSIM$\uparrow$ & PSNR$\uparrow$ & SSIM$\uparrow$ & PSNR$\uparrow$ & SSIM$\uparrow$ & PSNR$\uparrow$ & SSIM$\uparrow$ & &\\ 
\midrule
Restormer~\cite{zamir2022restormer} & 34.99        & 0.9511         & 34.34        & 0.9484         & 32.73        & 0.9330         & 31.35        & 0.9186        & 26.1        & 0.4406\\ 
Restormer*                          & 39.51        & 0.9756         & 39.87        & 0.9788         & 39.05        & 0.9760         & 38.32        & 0.9732        & 26.1        & 0.4417\\ 
Turtle~\cite{ghasemabadi2024lthm}   & 35.15        & 0.9467         & 35.38        & 0.9575         & 33.89        & 0.9456         & 32.55        & 0.9342        & 59.1        & 0.5514\\ 
Turtle*                             & 39.37        & 0.9765         & 39.35        & 0.9777         & 38.52        & 0.9745         & 37.73        & 0.9713        & 59.1        & 0.5529\\ 
STCNet~\cite{yang2024stc}           & 40.07        & 0.9799         & 39.51        & 0.9788         & 38.44        & 0.9753         & 37.79        & 0.9723        & 16.4        & 0.5974\\ 
ELEDNet~\cite{kim2024realevent}     & 39.51        & 0.9762         & 39.94        & 0.9800         & 39.07        & 0.9771         & 38.36        & 0.9743        & 12.8        & 0.5296\\ 
EFNet~\cite{sun2022eventfusion}     & 41.29        & 0.9895         & 40.84        & 0.9887         & 40.03        & 0.9865         & 39.37        & 0.9847        & \textbf{8.5}& \textbf{0.3984}\\ 
CBRDM~\cite{meng2025diffusion}      & 31.48        & 0.9388         & 31.26        & 0.9362         & 31.70        & 0.9437         & 30.70        & 0.9307        & 166.2       & 538.5\\
\multirow{2}{*}{Ours}   &\multirow{2}{*}{\textbf{41.88}}
                        &\multirow{2}{*}{\textbf{0.9912}} 
                        &\multirow{2}{*}{\textbf{41.47}} 
                        &\multirow{2}{*}{\textbf{0.9905}} 
                        &\multirow{2}{*}{\textbf{40.72}} 
                        &\multirow{2}{*}{\textbf{0.9887}}
                        &\multirow{2}{*}{\textbf{40.12}} 
                        &\multirow{2}{*}{\textbf{0.9874}}             
                        &\multirow{2}{*}{13.9}         
                        & $0.6736_{\scriptstyle N=5}$ \\
&  &  &  &  &  &  &  &  &  & $1.448_{\scriptstyle N=11}$ \\
\bottomrule
\end{tabular}

\label{tab:comparison}
\end{table*}

\begin{figure*}[htbp]
    \centering{\includegraphics[width=0.9\linewidth]{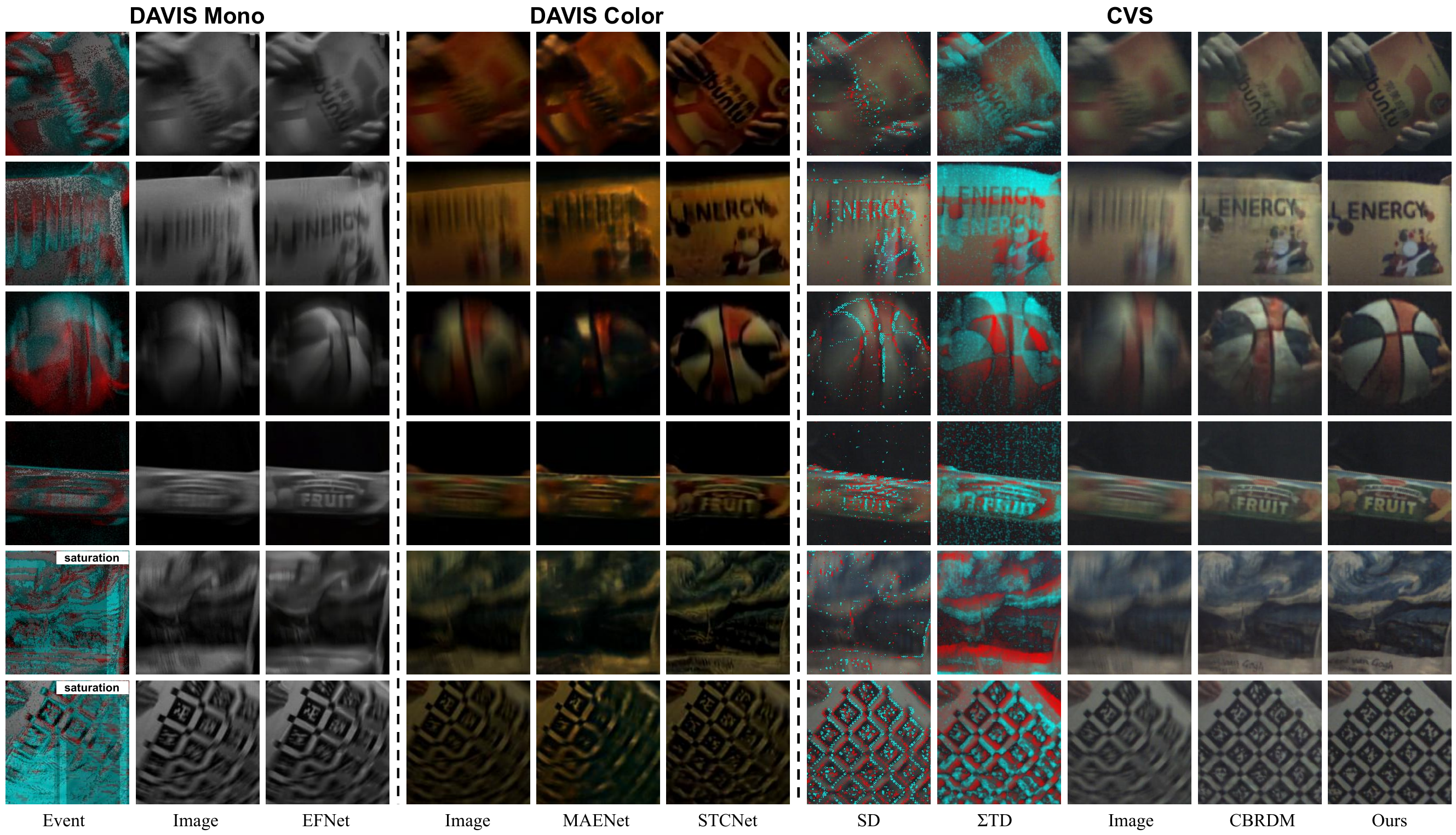}}\\
     \caption{Results on real-captured data compared with event-based methods and CVS-based method. 
     }
    \label{fig:5}
\end{figure*}

\begin{figure*}[htbp]
    \centering{\includegraphics[width=0.9 \linewidth]{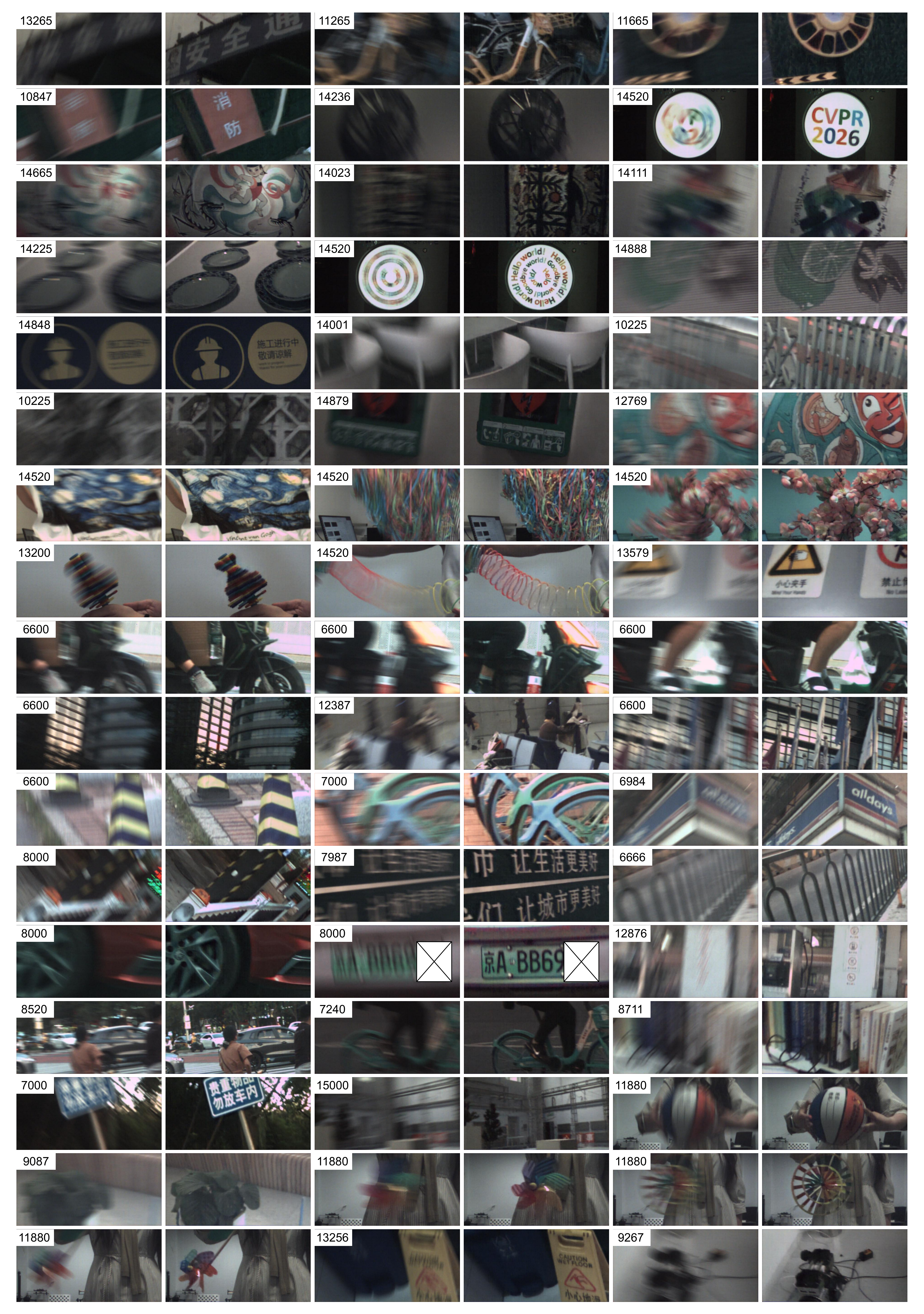}}\\
     \caption{Real-world deblurring results of CVS under different RGB exposure times (us).
     }
    \label{fig:4}
\end{figure*}

\section{Experimental Analysis}

\subsection{Datasets} \label{sec:dataset}


Due to the unique data modalities of different vision sensors, constructing a large-scale and highly generalizable deblurring dataset is essential for real-world deployment. Different from traditional software simulation methods, we reference a Digital Light Processing (DLP)-based vision chip characterization method~\cite{meng2025technicalreportdmdbasedcharacterization}. We employ a Digital Micro-Mirror Device (DMD) chip and its corresponding optical path to project light onto the sensor, similar to~\cite{onn_dmd}, thereby forming realistic sensor response that naturally includes non-ideal factors such as noise and nonlinearities. Hardware-level triggers ensure precise temporal synchronization between the DMD and the sensor, while fixed optical components guarantee pixel-level spatial alignment. This setup enables the conversion of high-frame-rate RGB datasets into modality-specific datasets for various vision sensors.

Specifically, we convert the SportsSloMo~\cite{SportsSlomo} dataset into the CVS format, obtaining the SportsSloMo-CVS dataset. 
The conversion process is summarized as follows. The DMD-based light projector sequentially projects the sharp images from SportsSloMo onto the CVS sensor, with one sharp frame projected during each $\tau_{\text{diff}} = 1{,}320$~µs interval.
Through its spatio-temporal difference pathways, the CVS produces an $\mathcal{SD}$ for every projected frame and a $\mathcal{TD}$ between each pair of consecutive frames. The RGB exposure duration of the CVS is configured to one of four settings: 6{,}600~µs, 9{,}240~µs, 11{,}880~µs, and 14{,}520~µs, corresponding to overlapping $N=5,7,9,11$ consecutive sharp RGB frames during projection. This procedure yields realistically blurred RGB images together with the corresponding $\mathcal{SD}$ and $\mathcal{TD}$ signals recorded within each exposure period. To obtain CVS ground-truth sharp frames, we project a single fixed sharp image throughout the entire RGB exposure duration, allowing the CVS to generate authentic RGB response. Ultimately, we obtain 98,569 training pairs, 1,928 validation pairs, and 1,820 test pairs, resulting in a large-scale, real-captured, pixel-level aligned, and scene-diverse dataset for multi-exposure and multi-form motion deblurring tasks with CVS.

\subsection{Quantitative Experiment on Synthetic Dataset‌}
To comprehensively evaluate the effectiveness of the proposed method, we compare with several representative deblurring approaches, including RGB-based methods \textit{Restormer}~\cite{zamir2022restormer} and \textit{Turtle}~\cite{ghasemabadi2024lthm}, CVS-based methods \textit{CBRDM}~\cite{meng2025diffusion}, as well as event-based methods \textit{EFNet}~\cite{sun2022eventfusion}, \textit{STCNet}~\cite{yang2024stc} and \textit{ELEDNet}~\cite{kim2024realevent}. 
Since the CVS provides spatio-temporal difference signals rather than events, we~\textbf{retrain} and evaluate all compared methods using the SportsSloMo-CVS dataset and their official implementations, with all models optimized on the same dataset for 10 epochs. 
For RGB-only methods, we conduct two experiments: (1) using only the blurry RGB input, and (2) extending the input modalities by concatenating the original three-channel RGB input with the $\mathcal{TD}$ and $\mathcal{SD}$ data along the channel dimension (denoted as *). 
For event-based methods, we replace the original event inputs with $\mathcal{TD}$ and $\mathcal{SD}$ data in the same configuration as ours. We employ CVS $\mathcal{TD}$ as a replacement for raw event inputs, grounded in the observation that despite the diverse event representations used in existing methods (e.g., voxels in STCNet/ELEDNet and SCER in EFNet), they share a common pipeline: accumulating $\pm$1-bit events into multi-bit values, partitioned into $C$ temporal bins, and formatted as $C{\times}H{\times}W$ tensors. Concatenating multi-frame multi-bit $\mathcal{TD}$ along the time dimension can approximate voxel representation.
Since the number of $\mathcal{TD}$ frames varies with the exposure duration, we fix non-CVS-based methods to 10 $\mathcal{TD}$ channels by using the available slices and zero-padding the rest, ensuring shape consistency without adding any extra information. Apart from this input-channel adjustment, all comparison architectures remain unchanged.
Considering that \textit{Turtle}, \textit{CBRDM}, \textit{STCNet}, and \textit{ELEDNet} require adjacent frames for prediction, we follow their usage protocols and uniformly exclude the boundary frames during testing to ensure evaluation on exactly the same set of samples.

We employ Peak Signal-to-Noise Ratio (PSNR) and Structural Similarity Index (SSIM) as evaluation metrics, with the results summarized in Table ~\ref{tab:comparison}. It can be observed that as the exposure time increases, the degree of motion blur in the input images intensifies, leading to a consistent performance drop across all methods. Nevertheless, under all four exposure durations ($N=5,7,9,11$), our method achieves the highest PSNR and SSIM scores.

Furthermore, Fig.~\ref{fig:3} presents qualitative comparisons among all methods. As can be seen, \textit{Restormer} and \textit{Turtle}, when relying solely on RGB input, tend to suffer from noticeable detail loss and structural blurring under severe motion blur. Notably, when incorporating $\mathcal{TD}$ and $\mathcal{SD}$ data, both methods achieve noticeable gains in detail recovery and structure sharpness. 
The diffusion-based CVS reconstruction method \textit{CBRDM} shows low quantitative accuracy and produces unrealistic structures and colors in the visualization results.
While event-based deblur methods such as \textit{STCNet}, \textit{ELEDNet}, and \textit{EFNet} can preserve more motion-related information in certain scenarios, they still suffer from varying degrees of artifacts, color distortions, and edge ghosting. In contrast, our method delivers sharper results with clearer edges and more accurate colors.

\subsection{Qualitative Comparisons on Real-World Data} \label{sec:compare_dvs}

To assess the real-world generalization of different sensor–algorithm combinations, we compare our method with other CVS-based approaches as well as event-camera-based pipelines using state-of-the-art deblurring algorithms.
As the CVS is a hybrid sensor that captures both RGB frames and spatio-temporal difference modalities, we select the RGB-event hybrid camera \textit{DAVIS} for a fair comparison. Considering that some compared methods are trained on grayscale inputs, we include both \textit{DAVIS Mono} (grayscale frames) and \textit{DAVIS Color} (color frames) in our experiments. DAVIS data is recorded using the official DV software with default threshold settings. The CVS and the DAVIS cameras are aligned to ensure overlapping fields of view and software synchronously record the same dynamic scenes, with the RGB/grey frame exposure time uniformly set to 14{,}520~µs.
The compared deblurring methods include CVS-based \textit{CBRDM}~\cite{meng2025diffusion} and event-based \textit{EFNet}~\cite{sun2022eventfusion}, \textit{STCNet}~\cite{yang2024stc}, \textit{MAENet}~\cite{Sun_2024_ECCV}. All models are evaluated using their officially released pretrained weights.
For methods offering multiple pretrained variants, we test all available options and report the best-performing results in Fig.~\ref{fig:5}. 
Our captured scenes span diverse fast-motion patterns, including rapid translation, 2D rotation, 3D sphere rotation, and irregular fabric motion.
Visualization results show that \textit{CBRDM} roughly restores structure but introduces noticeable color artifacts. Event cameras suffer from event-rate saturation under rapid motion, causing missing information, and even with valid events, several event-based methods still exhibit incomplete deblurring and color blending.
In contrast, our method preserves higher color fidelity and structural detail in real-world high-speed motion scenes.

\subsection{Real World Generalization Results}
Our model generalizes well to arbitrary exposure durations in real-captured data across diverse indoor and outdoor scenarios (Fig.~\ref{fig:4}; see Sec.~\ref{sec:enhance} for details on exposure time generalization).
The restored results remain sharp and color-consistent under varying motions and scenes.

\subsection{Ablation Study}

\begin{table}[htbp]
\small
\centering
\caption{Ablation study on different components.}
\begin{tabular}{c c c c|c c}
\toprule
$\mathcal{SD}$ & $\mathcal{TD}$ & CCF & TRRM & PSNR$\uparrow$ & SSIM$\uparrow$ \\ 
\midrule
$\times$ & $\times$ & $\times$ & $\times$ & 31.06 & 0.9429 \\ 
$\checkmark$ & $\times$ & $\checkmark$ &  $\times$ & 37.70 & 0.9811 \\ 
$\times$ & $\checkmark$ & $\checkmark$ &  $\times$ & 39.01 & 0.9842 \\ 
$\checkmark$ & $\checkmark$ & $\checkmark$ &  $\times$ & 39.45 & 0.9855 \\ 
$\checkmark$ & $\checkmark$ & $\times$ & $\times$ & 39.01 & 0.9841 \\
$\checkmark$ & $\checkmark$ & $\checkmark$ & $\checkmark$ & \textbf{40.12} & \textbf{0.9874} \\ 
\bottomrule
\end{tabular}

\label{tab:ablation}
\end{table}

We conduct ablation studies to quantify the contribution of each input modality and component of STGDNet. All experiments are performed on the $N{=}11$ test set with PSNR and SSIM as evaluation metrics, summarized in Table ~\ref{tab:ablation}.
\paragraph{Effect of Input Modalities.} We examine four input configurations: (1) RGB only, (2) RGB+$\mathcal{SD}$, (3) RGB+$\mathcal{TD}$, and (4) RGB+$\mathcal{SD}$+$\mathcal{TD}$. Compared to RGB-only input, incorporating $\mathcal{SD}$ improves PSNR by 6.64~dB and SSIM by 4.05\%, while adding $\mathcal{TD}$ achieves +7.95~dB and +4.38\%. Combining both modalities yields the best performance (+8.39~dB PSNR, +4.52\% SSIM), demonstrating that spatial and temporal difference cues jointly provide strong complementary motion information for deblurring.


\paragraph{Effect of Network Components.} We further evaluate the role of two key network modules: (1) the Temporal Recurrent Refinement Module; and (2) the Cross-modal Complementary Fusion. Results show that removing the TRRM (replaced by single forward pass) causes a drop of 0.67~dB in PSNR and 0.19\% in SSIM, leading to noticeable degradation in fine details and increased blurring of motion boundaries. Similarly, removing the CCF (replaced by directly concat with a two-layer convolution) results in a PSNR decrease of 0.44~dB and an SSIM reduction of 0.14\%. Taken together, these results confirm that both modules are essential for fully exploiting spatio-temporal difference data and leveraging the complementary nature of multi-modal features, making them indispensable components for high-quality motion deblurring.


\subsection{Performance Boundary Analysis}

\begin{figure}
    \centering{\includegraphics[width=0.9\linewidth]{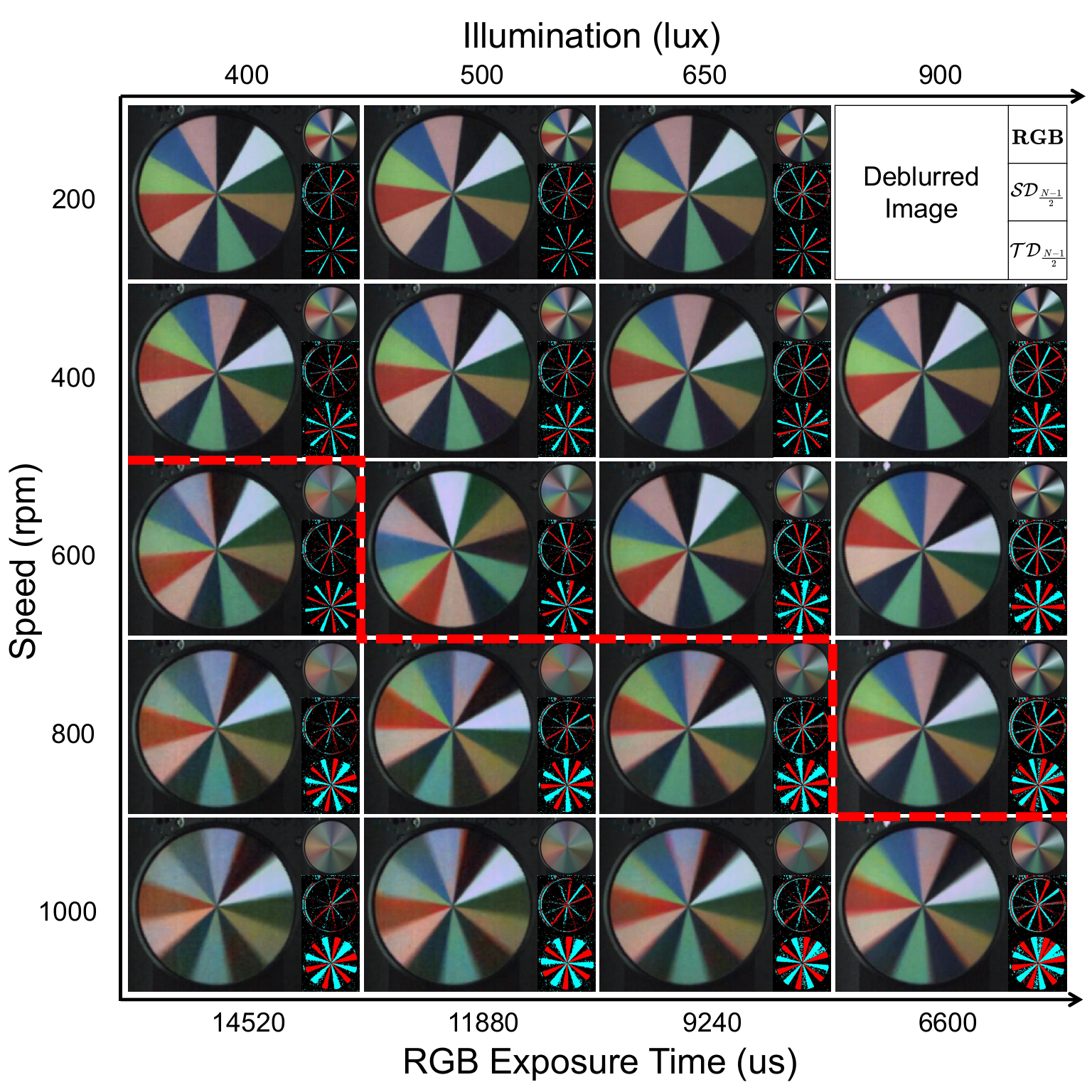}}\\
     \caption{Deblurring results across different rotational speeds and illumination levels in real-captured rotating-disk.
     }
    \label{fig:6}
\end{figure}

\begin{figure}
    \centering{\includegraphics[width=\linewidth]{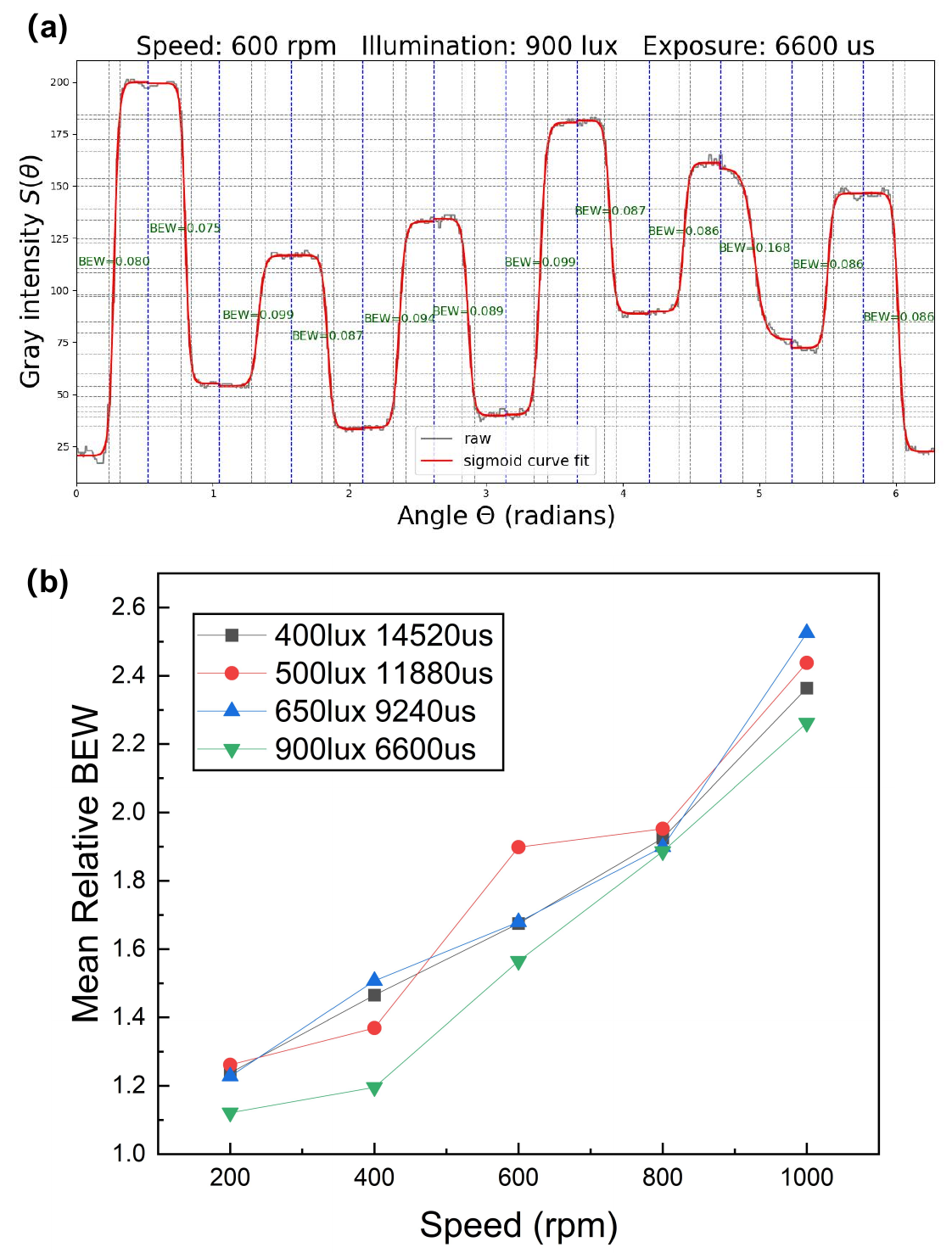}}\\
     \caption{Performance boundary visualization. (a) 1D angular intensity sequence sampled along the disk, together with the corresponding sigmoid fitting results, shown for one example configuration (600 rpm, 900 lux, $6{,}600$ µs exposure). (b) Mean Relative BEW versus rotation speed under different illumination levels.
     }
    \label{fig:7}
\end{figure}



To evaluate the performance limit of CVS-based deblurring in the real world, we establish a standardized rotating-disk benchmark that enables controlled, repeatable, and quantitatively measurable testing. The setup allows for adjustment of two key factors that  determine motion blur: disk rotation speed and RGB exposure duration.

As shown in Fig.~\ref{fig:6}, this benchmark reveals an approximate separating boundary in the rotation-speed–exposure-time plane: one region yields sharp and stable reconstructions, whereas the other leads to color mixing and algorithmic breakdown. Since ground-truth sharp images are unavailable in real-captured experiments, we follow the evaluation procedure of~\cite{dinh2023evaluation} and measure post-deblurring edge sharpness to obtain a quantitative indicator of performance trends. 

Specifically, we sample intensity profiles along the disk perimeter and model the transitions between adjacent sectors using a sigmoid function:
\begin{equation}
    S(\theta) = 
    \frac{\Delta}{1 + e^{-(a\theta + b)}} + g_{\min},
\end{equation}
where $\theta$ denotes the angular coordinate, $\Delta = g_{\max} - g_{\min}$ is the intensity range, and $(a,b)$ are the parameters controlling the slope and location of the transition. This formulation provides a smooth approximation of the intensity transition across the edge, as illustrated in Fig.~\ref{fig:7}(a).

Based on the fitted curves, we measure the blurred edge width (BEW) as the angular interval between the $10\%$ and $90\%$ intensity levels, and further normalize it with respect to a static reference image, then average over all edges to obtain the mean relative BEW (Mean-rBEW), which serves as a quantitative indicator of residual blur.

As shown in Fig.~\ref{fig:7}(b), Mean-rBEW increases with rotation speed, indicating that stronger motion leads to more severe residual blur.
In contrast, the Mean-rBEW curves under different exposure durations show no pronounced differences, suggesting that our method is robust to varying exposure conditions and illumination conditions.

\subsection{Enhancing Modeling Flexibility}
\label{sec:enhance}
We further investigate how to leverage the intrinsic properties of the $\mathcal{SD}$ and $\mathcal{TD}$ modalities to enable more flexible spatio-temporal modeling, without modifying the architecture of STGDNet.

\paragraph{Single-frame to Video.}

In our default formulation (Sec.~\ref{sec:problem}, Eq.~\ref{eq:problem_formulation}), the model takes the $\mathcal{SD}$ frame closest to the exposure midpoint, $\mathcal{SD}_{\lfloor (N-1)/2 \rfloor}$, as structural guidance, such that the restored image $\mathbf{D}$ is aligned with this temporal slice. 

Notably, this design can be naturally extended by selecting different $\mathcal{SD}$ frames within the exposure window. Specifically, the model can be generalized as:
\begin{equation}
\mathbf{D}_{k} = 
\mathcal{M}^{*}\!\left(\mathbf{B},\, \mathcal{SD}_{k},\, \{\mathcal{TD}_{i}\}_{i=0}^{N-2}\right),
\end{equation}
where the reconstructed image $\mathbf{D}_{k}$ is structurally aligned with the chosen $\mathcal{SD}_{k}$.

By training the network with the same procedure, $\mathcal{M}^{*}$ can learn to restore the blurry input $\mathbf{B}$ to arbitrary temporal slices within the exposure period, i.e., $\mathbf{D}_{k}$ for any $k \in [0, N-1]$. At inference time, this enables the recovery of a sequence of temporally consistent frames from a single blurry RGB image and its corresponding spatio-temporal difference signals, effectively reconstructing the scene dynamics within the exposure duration.

Additional video results are provided on the project page.

\paragraph{Exposure Time Generalization.}

According to Sec.~\ref{sec:problem}, the length of the temporal-difference sequence is determined by 
$N = \left\lceil \frac{t_{\text{RGB}}}{\tau_{\text{diff}}} \right\rceil$, 
which ensures full coverage of motion within the exposure period. However, when $t_{\text{RGB}}$ is not an integer multiple of $\tau_{\text{diff}}$, the last element $\mathcal{TD}_{N-2}$ may contain additional motion information beyond the actual exposure window.

To enable generalization to arbitrary continuous exposure times, we introduce a temporal-augmentation strategy during training. Specifically, we randomly extend the last temporal-difference entry by accumulating subsequent TD frames:
\[
\mathcal{TD}_{N-2}^\star = \sum_{j=0}^{m} \mathcal{TD}_{N-2+j},
\]
where $m \in \{1,2,3\}$ is randomly sampled. This simulates the situation where the final $\mathcal{TD}$ frame contains extra motion beyond the exposure period while requiring the network to learn to extract only the useful temporal cues.

We compare models trained with and without this strategy on discrete exposure settings. As shown in Table~\ref{tab:td_aug}, the augmentation introduces negligible performance differences, while significantly improving generalization to continuous exposure durations. This indicates that the augmented model successfully learns to adaptively select valid temporal information, enabling generalization to arbitrary continuous exposure durations.

\begin{table}[htbp]
\centering
\caption{Effect of TD-sequence augmentation on deblurring performance.}
\scriptsize
\setlength{\tabcolsep}{3pt}
\begin{tabular}{c|p{0.7cm} p{0.7cm}|p{0.7cm} p{0.7cm}|p{0.7cm} p{0.7cm}|p{0.7cm} p{0.7cm}}
\toprule
\multirow{2}{*}{Augment} 
& \multicolumn{2}{c|}{5}  
& \multicolumn{2}{c|}{7}   
& \multicolumn{2}{c|}{9}   
& \multicolumn{2}{c}{11} 
\\
& PSNR & SSIM 
& PSNR & SSIM 
& PSNR & SSIM 
& PSNR & SSIM 
\\
\midrule
$\times$     
& 41.88 & 0.9912
& 41.47 & 0.9905
& 40.72 & 0.9887
& 40.12 & 0.9874
\\
$\checkmark$       
& 41.88 & 0.9913
& 41.46 & 0.9905
& 40.68 & 0.9887
& 40.05 & 0.9873
\\
\bottomrule
\end{tabular}
\label{tab:td_aug}
\end{table}
\section{Conclusion}
We present a CVS-based motion deblurring framework that leverages high-speed spatio-temporal difference signals to guide RGB deblurring.  
Built upon the hardware-level disentanglement of structural and motion cues, our STGDNet recurrently fuses these complementary modalities through multi-branch cross-attention, enabling accurate recovery of sharp and color-consistent images under extreme motion.  
Extensive experiments show that our method outperforms existing state-of-the-art methods and generalizes well across diverse environments, revealing new opportunities for high-fidelity motion deblurring with CVS.


\begingroup
\renewcommand\section*{} 
\noindent\textbf{Acknowledgements.} This work was supported by the National Key Research and Development Program of China (No. 2025YFG0100200) and the Tsinghua University Initiative Scientific Research Program 20257020014.
\endgroup

{
    \small
    \bibliographystyle{ieeenat_fullname}
    \bibliography{main}
}



\end{document}